\documentclass[lettersize,journal]{IEEEtran}
\usepackage{amsmath,amsfonts}
\usepackage{amsthm}
\usepackage{amssymb}
\usepackage[ruled,linesnumbered]{algorithm2e}
\usepackage{array}
\usepackage{bm}
\usepackage{booktabs}
\usepackage{textcomp}
\usepackage{stfloats}
\usepackage{url}
\usepackage{verbatim}
\usepackage{graphicx}
\usepackage{cite}
\usepackage{multirow}
\usepackage{threeparttable}
\usepackage{flushend}

\usepackage{caption}
\usepackage{pifont}
\usepackage{subcaption}
\usepackage{color}
\usepackage{xcolor}
\usepackage{colortbl,booktabs}
\usepackage[pagebackref,breaklinks,colorlinks]{hyperref}

\definecolor{tablered}{RGB}{205,51,51}
\definecolor{tablegreen}{HTML}{39b54a}
\definecolor{tableblue}{HTML}{4682B4}

\definecolor{figgrey}{RGB}{128,128,128}
\definecolor{figcyan}{RGB}{0,139,139}
\definecolor{figdarkred}{RGB}{64,0,0}
\definecolor{figred}{RGB}{205,51,51}
\definecolor{figgreen}{RGB}{113,198,113}

\newcommand{\eg}{\textit{e}.\textit{g}.}

\begin{document}

\title{Image Enhancement for Remote Photoplethysmography in a Low-Light Environment}

\author{\parbox{16cm}{\centering
		{\large Lin Xi$^1$, Weihai Chen$^1$, Changchen Zhao$^{*2}$, Xingming Wu$^1$, and Jianhua Wang$^1$}\\
		{\normalsize
			$^1$ School of Automation Science and Electrical Engineering, Beihang University, Beijing 100191, China\\
			$^2$ College of Information Engineering, Zhejiang University of Technology, Hangzhou 310023, China \\
			xilin1991@buaa.edu.cn, whchenbuaa@126.com, cczhao@zjut.edu.cn, wxmbuaa@163.com, jhwang@126.com}}
	\thanks{Correspnding Author: Changchen Zhao, email: cczhao@zjut.edu.cn}
}



\maketitle

\begin{abstract}
	With the improvement of sensor technology and significant algorithmic advances, the accuracy of remote heart rate monitoring technology has been significantly improved. Despite of the significant algorithmic advances, the performance of rPPG algorithm can degrade in the long-term, high-intensity continuous work occurred in evenings or insufficient light environments. One of the main challenges is that the lost facial details and low contrast cause the failure of detection and tracking. Also, insufficient lighting in video capturing hurts the quality of physiological signal. In this paper, we collect a large-scale dataset that was designed for remote heart rate estimation recorded with various illumination variations to evaluate the performance of the rPPG algorithm (Green, ICA, and POS). We also propose a low-light enhancement solution (technical solution) for remote heart rate estimation under the low-light condition. Using collected dataset, we found 1) face detection algorithm cannot detect faces in video captured in low light conditions; 2) A decrease in the amplitude of the pulsatile signal will lead to the noise signal to be in the dominant position; and 3) the chrominance-based method suffers from the limitation in the assumption about skin-tone will not hold, and Green and ICA method receive less influence than POS in dark illuminance environment. The proposed solution for rPPG process is effective to detect and improve the signal-to-noise ratio and precision of the pulsatile signal. Our dataset is available at \href{https://github.com/xilin1991/Large-scale-Multi-illumination-HR-Database}{https://github.com/xilin1991/Large-scale-Multi-illumination-HR-Database}.
\end{abstract}


\section{INTRODUCTION}
\label{sec:Intro}

The advances in semiconductor technology enables us to measure one's physiological parameters in a non-contact way just by means of simple instrumentations such as consumer-level digital camera \cite{Verkruysse1, Poh1, Poh2} and cellphone \cite{Pelegris}. The technology is usually called remote photoplethysmography (rPPG) or image plethysmography (iPPG). The non-contact rPPG does not need any direct physical contact with the subject, which introduces great convenience in numerous applications such as driving monitoring, fitness monitoring\cite{Lin2018StepCA, Wang2017RobustHR}, home healthcare \cite{HomeHealth, FVP}, face anti-spoofing \cite{anti1}, etc.

However, there exists one question that is crucial but has not gained much attention: the impact of insufficient illuminance on the rPPG measurement. For the experimental environment, the performance of the existing approach is verified under well-lit conditions and there are no datasets acquired under various illuminance to compare the accuracy of the existing rPPG algorithm. As a consequence, monitoring heart rate during activities that occur in evenings or insufficient light environments, such as sleeping, driving train or car, etc., can remain a new challenge. But in the past, the researcher put less attention to this issue and done not some work.

The challenges of extracting physiological signal in low light environment are threefold. First, the region of interest (ROI) cannot be reliably detected. Most object detection and tracking algorithms determine the target based on rich features contained in the images, \eg, Scale-Invariant Feature Transform (SIFT) \cite{SIFT}, Histogram of Gradient (HOG) \cite{HOG}, or Convolutional Neural Network (CNN) based features \cite{CNN}. However, insufficient illuminance attenuates the light intensity of the scene, resulting in not enough features inside the target for the trackers or low color contrast between the target and the background. Second, the physiological signal captured by the camera is extremely subtle and this problem is more significant in low-lighting conditions. A decrease in the amplitude of the pulsatile signal will lead to the noise signal to be in the dominant position. Third, the pulse extraction models may fail because their assumptions no longer hold. For example, the skin reflection model \cite{Wang_POS, Haan2} relies heavily on the skin-tone direction and the pulsatile color variation direction in RBG space. The directions will change in low light condition because the signals in blue and red channels are almost zeros.

In this paper, we propose a novel framework to extract rPPG pulse information in low light condition. Inspired by recent progress in low-light image enhancement techniques \cite{MSRCR, Naturalness, LIME}, we employ image enhancement algorithm prior to pulse extraction for accurate heart rate measurement such that the physiological signals would not be influenced by illumination and existing rPPG methods can be combined directly with the processing pipeline. The solution consists of three main steps: 1) video enhancement; 2) ROI detection and tracking; and 3) heart rate extraction.

The main contributions of this article are summarized as follows:

1)	To the best of our knowledge, we provide the first solution for robust rPPG measurement under low light environment, which consists of an image enhancement module and a pulse extraction module.

2)	We are the first to systematically analyze the light illumination on the rPPG measurement. The accuracy of several rPPG algorithms are tested on a wide variety of light illuminance in terms of mean absolute error (MAE), root mean squared error (RMSE) and signal-to-noise ratio (SNR) \cite{Haan2}.

3)	We build large-scale dataset with the illuminance covering a wide range from 1 to 100 lux. To the best of our knowledge, there is no publicly available datasets for the research of rPPG measurement under low light environment.

The rest of this article is organized as follow: Section \ref{sec:RW} introduces the related works of existing publicly available remote heart rate datasets and remote heart rate estimation. Section \ref{sec:Met} presents a framework of low-light video enhancement for remote heart rate measurement in details. Experimental setup and results are reported in Section \ref{sec:Exp} and \ref{sec:Res}, respectively. Section \ref{sec:Con} concludes the whole article.

\section{RELATED WORK}
\label{sec:RW}

\subsection{Publicly Available Datasets for Remote Heart Rate Estimation}
\label{subsec:RW_dataset}

Many of the published methods conducted the experiment on private datasets, which cause difficulties in fair comparisons of different methods. But there are a lot of publicly available datasets that were designed for rPPG methods. One of them is the COHFACE dataset \cite{dataset1}, which is composed of 160 videos and physiological signals collected from 40 healthy individuals. The environment on the scene was changed once as to create two types of lighting conditions, studio and natural. At the same time, Tulyakov et al. introduced a new dataset MMSE-HR \cite{dataset2} , which was part of the MMSE database \cite{MMSE} and the subjects` facial expressions were more varied. Besides, the MAHNOB-HCI dataset \cite{dataset3} was introduced by Li et al \cite{Li_CVPR}. for evaluating their method performance. However, the datasets mentioned above were either designed for other application or published earlier.

\begin{table*}[!htb]
	\caption{A summary of publicly available datasets for remote heart rate estimation}
	\begin{center}
		\resizebox{1.0\textwidth}{!}{
			\begin{tabular}{|c||c||c||c||c||c|||c||c|}
				\hline
				\textbf{Name} & \textbf{Date published} & \textbf{Illumination} & \textbf{Subject} & \textbf{Recording devices} & \textbf{Number of videos} & \textbf{Size} & \textbf{Total duration} \\ \hline
				MAHNOB-HCI \cite{dataset3} & 2012 & Normal & 27 & RGB camera (60fps) & 522 & 25.2 GB & 527 min\\
				COHFACE \cite{dataset1} & 2016 & Normal & 40 & RGB camera (20fps) & 160 & 368 MB & 160 min\\
				MMSE-HR \cite{dataset2} & 2016 & Normal & 40 & RBG camera (25fps) & 102 & - & - \\
				OBF \cite{Xiaobai} & 2018 & Normal & 106 & RGB camera (60fps)/NIR camera (30fps) & 212 & - & 1060 min \\
				PBD-RPPG \cite{Hoffman_Lakens_2019} & 2018 & 3 level illuminance & 3 & RBG camera & 21 & 16.8 GB & 63 min\\
				VIPL-HR \cite{Niu2018VIPLHRAM} & 2018 & Dim/Bright illuminance & 107 & RGB camera (25fps)/NIR camera (30fps) & 3203 & 48 GB (compressed) & 1189 min\\ \hline
			\end{tabular}
		}
	\end{center}
	\label{tab:sum}
\end{table*}

In the last two years, there are also a few publicly available datasets specially designed for the task of remote heart rate estimation. In 2018, Xiaobai et al. proposed a database designed for HR and heart rate variability (HRV) measurement \cite{Xiaobai}. Since this database aims at HRV analysis and all the situations in this database are well-controlled, making it very easy for remote HR estimation. Hoffman et al. released a public benchmark dataset for testing rPPG algorithm performance (PBD-RPPG) \cite{Hoffman_Lakens_2019}. In the datasets, Videos accompanied by ECG measurements were recorded on three participants with three different skin tones. Niu et al. introduced a large-scale multi-modal HR dataset (named as VIPL-HR) \cite{Niu2018VIPLHRAM}, which contains 2451 visible light videos and 752 near-infrared videos of 107 subjects. As we can see from Table \ref{tab:sum}, the existing public-domain datasets for remote heart rate estimation are limited in the normal light environment.

\subsection{Remote Heart Rate Estimation}
\label{subsec:RW_HR}

Over the last decades, researchers have been investigating the feasibility of remote heart rate estimation using a non-contact-based device such as a camera. Pioneering research includes Verkruysse et al. \cite{Verkruysse1} detect PPG signals captured by regular color digital cameras. This research reveals the green light (approximately 530 nm wavelength) has been found to give superior results compared to red or blue light \cite{Verkruysse1, Matsumura} and also compared to infra-red \cite{Maeda}. Poh et al. applied BSS-based model to remote heart rate estimation \cite{Poh1}. They separate the PPG signal from the red, green, and blue color channel in the video via Independent Component Analysis (ICA) \cite{Poh1}, and then Fourier transformation was applied. Another kind of rPPG algorithm is based on the skin reflection model \cite{Wang_POS} which exploits the interaction between light and skin. The CHROM \cite{Haan2} uses a linear combination of the chrominance signals by assuming a fixed skin-tone to white-balance the image. In the later work of \cite{Haan2}, Wang et al. comprehensively investigates the algorithmic principles of rPPG in a mathematical context with optical and physiological reasoning and proposes a new method, Plane-Orthogonal-to-Skin (POS) \cite{Wang_POS} , which project original RGB signal to extract pulsatile signals for remote heart rate estimation.

This article investigates the robustness of remote heart rate estimate algorithms (Green, ICA, POS) of under various illumination conditions, especially in the low-light environment, further study is carried out to assess their performance under various light intensity.

\section{METHOD}
\label{sec:Met}

\begin{figure*}
	\begin{center}
		\includegraphics[width=1.0\linewidth]{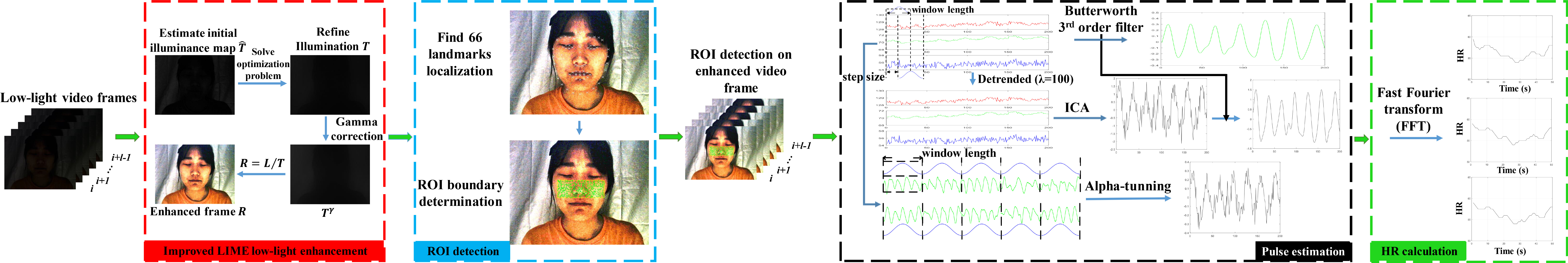}
	\end{center}
	\caption{Overview of the improved low-light video enhancement solution for rPPG.}
	\label{fig:flowchart}
\end{figure*}

We propose a solution for remote heart rate estimation under low-light condition, which consists of three stages: low-light enhancement, ROI detection and tracking, and remote heart rate estimation. The flowchart is shown in Fig. \ref{fig:flowchart}. The purpose of the first stage is to enhance the input video frame where visibility of the physiological signal is improved. The output is an enhanced RGB image for each frame in the video sequence. Then the ROI that contains the pulsatile signal will be obtained from enhanced video frames in the second stage. The last stage is heart rate extraction from raw traces in the ROI.

\subsection{Low-light Enhancement}
\label{subsec:Met_LL}

Regular remote heart rate estimation methods perform well in normal light environment, but the regular pipeline may eventually lead to failure if the videos are recorded under more challenging conditions, \eg, low-light illumination. Inspired by Retinex model \cite{Retinex1, Retinex2}, we can enhance the video in the low-light environment via removing the impact of illumination to facilitate the process of remote heart rate estimation. Referring to Retinex model, the frame of low-light video is decomposed into reflectance and illumination, which can be modeled as:
\begin{equation}
	\textbf{L} = \textbf{R} \circ \textbf{T}
	\label{con:Retinex}
\end{equation}
where $\textbf{L}$ is the frame in the video, $\textbf{R}$ and $\textbf{T}$ represent the reflectance and the illumination map, respectively. The operator $ \circ$ denotes the element-wise multiplication. By transforming (\ref{con:Retinex}), we can recover the enhanced result $\textbf{R}$ by the division is element-wise and $\textbf{T}$ is the key to recover the enhanced image. We observe that the green channel contains the strongest pulsatile signal among the RGB color channels under low-light condition. Our initial illumination map adopts the value of the green channel to prevent mixing of noise from other channels. The operation is as follows:
\begin{equation}
	\widehat{\textbf{T}}\leftarrow \textbf{L}^{G}
\end{equation}
for each individual pixel, where G denotes green channels in the three color channels.

In this work, we aim to preserve the over-all structure, smooth the textural details, and amplify the pulsatile signal. To this end, based on our employed initial illumination map $\widehat{\textbf{T}}$, we solve the following optimization problem, which is proposed in LIME \cite{LIME},
\begin{equation}
	\mathop{\min}_\textbf{T} \parallel{\widehat{\textbf{T}}} - \textbf{T} \parallel^2_\textbf{\textit{F}} + \alpha\parallel\textbf{W} \circ \nabla\textbf{T}\parallel_\textbf{1}
	\label{eq: optimization}
\end{equation}
where $\parallel\cdot\parallel_\textbf{\textit{F}}$ and $\parallel\cdot\parallel_\textbf{1}$ denote the Frobenious and $ \ell_1$ norms, respectively, $\alpha$ is the coefficient to balance the involved two terms, and $\textbf{W}$ is the weight matrix, and $\nabla\textbf{T}$ is the first order derivative filter. In Equation (\ref{eq: optimization}), initial map $\widehat{\textbf{T}}$ refers to green channel to preserve physiological signal.

\subsection{ROI Detection and Tracking}
\label{subsec:Met_ROI}

\begin{figure}[!htb]
	\begin{center}
		\includegraphics[width=0.9\linewidth]{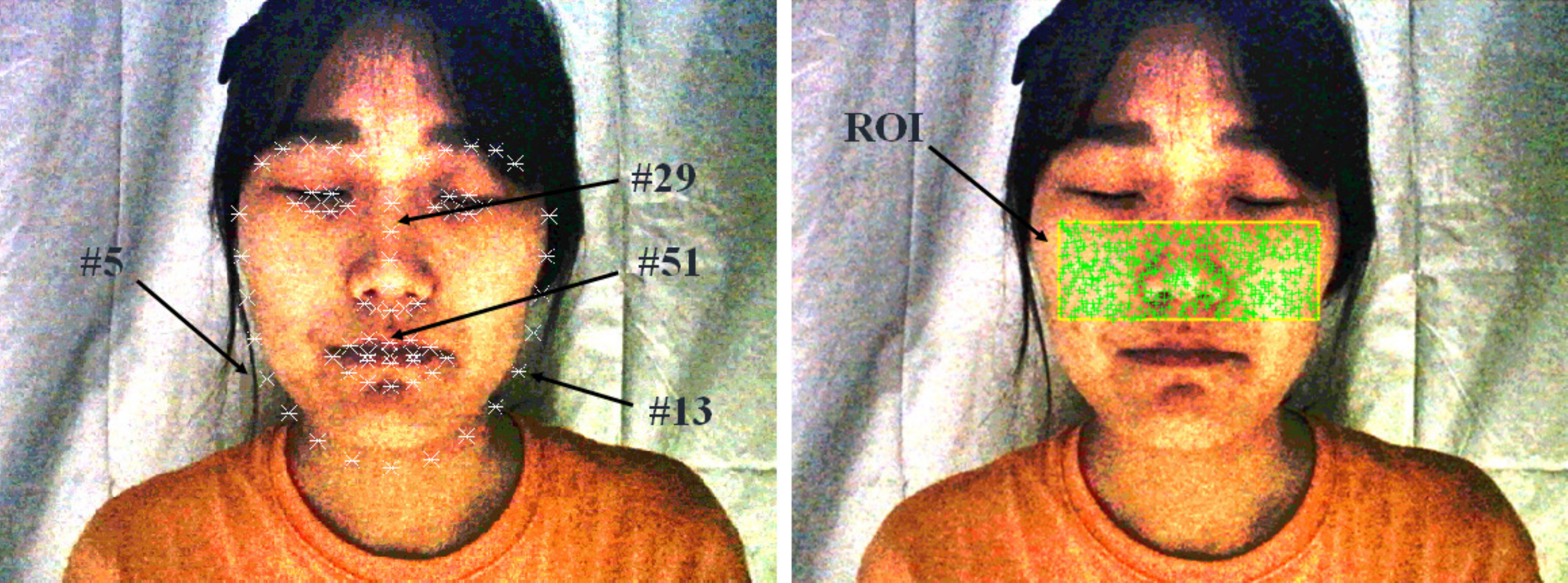}
	\end{center}
	\caption{ ROI detection and tracking. The white maker symbol $*$ indicate 66 facial landmarks and we select $n=5, 13, 29, 51$ points to define the initial ROI boundaries. The yellow line shows the ROI boundary, inside which feature points are detected and tracked.}
	\label{fig:ROI}
\end{figure}

The purpose of stage 2 is to obtain the ROI that contains the pulsatile signal from the enhanced frame of video. We first use the Viola-Jones detection algorithm \cite{Viola_Jones1, Viola_Jones2} to detect the location of a face on the first enhanced frame of the input video, then adopt Discriminative Response Map Fitting (DRMF) method \cite{Asthana} to find the coordinates of 66 facial landmarks inside the face region. We use $n=4$ points out of 66 landmarks to define our ROI and generate a rectangular mask of the ROI as the yellow bounding box shown in Fig. \ref{fig:ROI}.

The rules of defining the ROI is as follows: the first one is to exclude the mouth and eye region; the second one is that the left and right boundaries of the ROI do not exceed the cheek width. Therefore, we select $n=5, 13, 29, 51$ points to define initial ROI boundaries, and then indent the boundaries of ROI from initial boundaries above.

Then we use Staple tracker \cite{Bertinetto} to counter the problem of ROI movement for the subsequent frames. Given the ROI of frame $i$, spatial averaging \cite{Poh1} is applied to compute the average value of pixels in each color channel within the ROI. This results in the raw traces:
\begin{equation}
	\textbf{C(i)} = \left [ \vec{C}^R(i), \vec{C}^G(i), \vec{C}^B(i) \right ]^T
\end{equation}
for $i = 1,...,L$, where L is the number of frames in the video.

\subsection{Heart Rate Estimation}
\label{subsec:Met_HR}

Three published rPPG algorithms (Green, ICA and POS) are used to extract the blood volume pulse from the enhanced video to compare the performance of these algorithms under low-light conditions.

\textbf{Green} (Verkruysse, Svaasand \& Nelson, 2008 \cite{Verkruysse1}): In this case, we first select the green color channel vector $\vec{C}^G(i)$ from the raw traces. The signal, $\vec{C}^G(i)$, was selected as the channel with the greatest frequency power in the range 0.7 and 2.5 Hz.

\textbf{ICA} (Poh, McDuff \& Picard, 2010 \cite{Poh1}): By referring to priors approach \cite{iPhys}, we detrended the raw traces based on smoothness with $\lambda=100$. Then, ICA method is applied to the normalized detrended signal and filtered using a zero-phase, 3rd-order Butterworth bandpass filter with pass band frequencies of [0.7 2.5] Hz (corresponding [42 150] beat per minute, BPM).

\textbf{POS} (Wang, Brinker, Stuijk \& de Haan, 2017 \cite{Wang_POS}): In this method a projection plane orthogonal to the skin-tone is used for pulse extraction. A moving window of length 1.6 seconds was set for the raw traces. For each window of data, the raw traces data are normalized by their respective mean to give $ \widetilde{\textbf{C}}_n(i) \in \mathbb{R}^{3\times L_W} $, where $ L_W $ is the length of window. Next, the normalized color matrix, $ \widetilde{\textbf{C}}_n(i) $, was multiplied by the projection matrix $\textbf{P}$ to give $\textbf{S}(i)$. The resulting outputs of the window-based analysis was used to construct the blood volume pulse signal in overlap add fashion.

\section{EXPERIMENTAL SETUP AND IMPLEMENTATION DETAILS}
\label{sec:Exp}

This section presents the experimental setup for the benchmarking. First, a self-collected large video dataset recorded under various illumination conditions is introduced. Next, evaluation metrics are presented. Finally, we present the parameters of the experiment.

\subsection{Dataset}
\label{subsec:Exp_dataset}

We recruited 15 healthy subjects (12 male, 3 female, 18 to 30 years old) in this experiment and a total number of 165 video sequences were recorded under various illuminations. The dataset will be available publicly\footnote{https://github.com/xilin1991/Large-scale-Multi-illumination-HR-Database}. The detail of the experimental setup is described as follow:

\textbf{Apparatus}: Logitech HD pro webcam C930E color camera was used to record videos of $640 \times 480$ pixels, 30 fps, 60s, uncompressed bitmap format. PPG signal was measured using CONTEC CMS50E.

\begin{figure}[!htb]
	\begin{center}
		\includegraphics[width=1.0\linewidth]{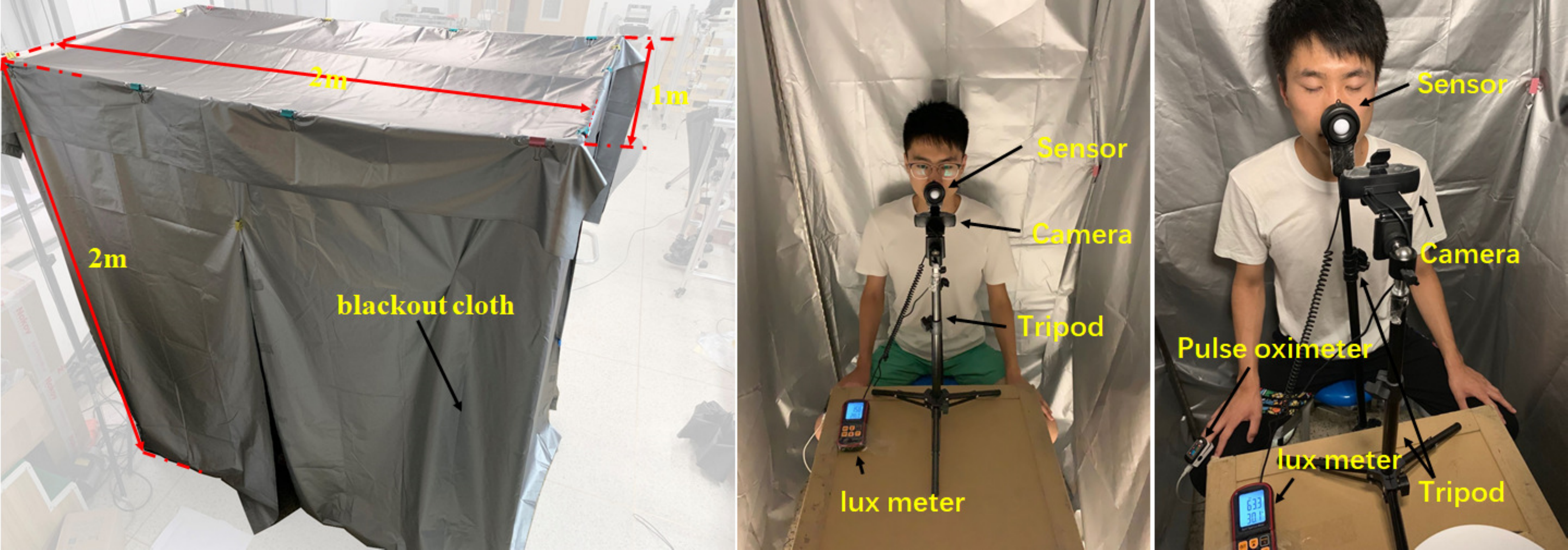}
	\end{center}
	\caption{The appearance of darkroom and experimental setup.}
	\label{fig:darkroom}
\end{figure}

\textbf{Experimental environment}: The video is captured in a darkroom as shown in Fig. \ref{fig:darkroom}, which is covered with a blackout cloth in order to isolate from ambient light. A split type illuminometer, BENETECH GM1030 lux meter, is used as the instrument for measuring light intensity and brightness in the darkroom.

\textbf{Participants}: Fifteen participates were recruited in our experiment. In the first step of the experiment, we measure the light intensity received in the region of face via the illuminometer which is placed against the light source and was positioned in front of the participant`s face. During the experiment, each of the 15 individuals was asked to sit 1 meter away from the camera, and the camera was positioned at the participant`s nose level as shown in Fig. \ref{fig:darkroom}.

\subsection{Evaluation Metrics}
\label{subsec:Exp_Eval}

\textbf{Signal-to-Noise-Ratio (SNR)} is adopted from \cite{Haan2} to evaluate the quality of the extracted pulse signal in comparison with the ground truth signal. It is derived by the ratio of the first two harmonic ranges to the remaining parts of the pulse signal in the frequency domain. The valid frequency range considered in SNR is [42 240] (equivalent to [0.7 4] Hz).

\textbf{Heart rate accuracy}. We extract the instantaneous heart rate estimation from a set of 51 overlapping 10-second windows for each 1-minute video. The first 10-second windows in the video as initialization is used to calculate the tenth-second instantaneous heart rate. We choose mean absolute error (MAE) and root mean squared error (RMSE) to evaluate the accuracy of heart rate estimation.

\subsection{Parameters Determination}
\label{subsec:Exp_Para}

We chose to vary the illuminance for three compared rPPG methods (Green, ICA, POS) in order to understand the influence of light intensity on accuracy of pulse recovery. For completeness and anticipating effects on rPPG signal, we varied illuminance in the range: \{$10^0,\ 10^{0.2},\ 10^{0.4},\ 10^{0.6},\ 10^{0.8},\ 10^{1.0},\ 10^{1.2},\ 10^{1.4},\ 10^{1.6}\\,\ 10^{1.8},\ 10^{2.0}$\} lux (equivalent to \{1.0, 1.6, 2.5, 4.0, 6.3, 10.0, 15.8, 25.1, 39.8, 63.1, 100.0\} lux).

\begin{table*}[!htb]
	\caption{Summary of the BVP SNR, MAE and RMSE for the rPPG algorithms under various light intensity in original video and different enhancement screens.}
	\begin{center}
		\resizebox{0.8\textwidth}{!}{
			\begin{tabular}{|c||c|c||ccccccccccc|}
				\hline
				\multicolumn{3}{|c||}{Lux} & $10^{0.0}$ & $10^{0.2}$ & $10^{0.4}$ & $10^{0.6}$ & $10^{0.8}$ & $10^{1.0}$ & $10^{1.2}$ & $10^{1.4}$ & $10^{1.6}$ & $10^{1.8}$ & $10^{2.0}$             \\ \hline
				\multirow{12}{*}{\textbf{SNR(dB)}}   & \multirow{3}{*}{\textbf{Original}} & Green & -1.06          & -0.75          & 1.13          & 2.13          & 5.06          & 6.32          & 7.32          & 7.01          & 7.98           & 8.53           & 7.51           \\
				&                                    & ICA   & -0.50          & -0.08          & 1.18          & \textbf{2.72} & \textbf{6.03} & \textbf{6.61} & \textbf{7.83} & \textbf{8.98} & \textbf{10.15} & \textbf{12.33} & \textbf{11.65} \\
				&                                    & POS   & -2.9           & -2.54          & -1.01         & -0.87         & 2.59          & 4.72          & 6.00          & 7.17          & 8.23           & 9.49           & 9.20           \\ \cline{2-14}
				& \multirow{3}{*}{\textbf{HE}}       & Green & -1.37          & 0.41           & 0.30          & 1.24          & 3.55          & 4.32          & 6.07          & 5.17          & 5.65           & 5.45           & 4.15           \\
				&                                    & ICA   & -              & -              & -             & -             & -             & -             & -             & -             & -              & -              & -              \\
				&                                    & POS   & -2.63          & -1.93          & -1.26         & -0.93         & 2.62          & 4.56          & 5.59          & 6.81          & 8.10           & 9.18           & 9.33           \\ \cline{2-14}
				& \multirow{3}{*}{\textbf{LIME}}     & Green & \textbf{-0.19} & 0.03           & 1.00          & 0.11          & 1.45          & 3.72          & 5.38          & 6.61          & 6.63           & 9.46           & 6.70           \\
				&                                    & ICA   & -              & -              & -             & -             & -             & -             & -             & -             & -              & -              & -              \\
				&                                    & POS   & -2.44          & -1.77          & -1.25         & -0.91         & 2.23          & 4.64          & 4.54          & 6.59          & 6.66           & 9.17           & 7.40           \\ \cline{2-14}
				& \multirow{3}{*}{\textbf{Improved}} & Green & -0.26          & \textbf{1.57}  & \textbf{1.30} & 1.63          & 3.62          & 3.07          & 3.65          & 3.28          & 3.16           & 2.91           & 4.04           \\
				&                                    & ICA   & -              & -              & -             & -             & -             & -             & -             & -             & -              & -              & -              \\
				&                                    & POS   & -2.42          & -1.24          & -0.89         & 0.81          & 2.08          & 1.67          & 1.86          & 1.97          & 2.05           & 2.06           & 3.25           \\ \hline
				\multirow{12}{*}{\textbf{MAE(BPM)}}  & \multirow{3}{*}{\textbf{Original}} & Green & 14.93          & 14.06          & 7.87          & 5.35          & \textbf{2.45} & 2.24          & 2.35          & 2.41          & 1.72           & 1.47           & 1.76           \\
				&                                    & ICA   & 14.96          & 14.54          & 8.04          & 6.76          & 3.59          & 3.39          & 2.32          & 2.34          & 1.80           & 1.17           & 1.69           \\
				&                                    & POS   & 59.89          & 51.38          & 29.37         & 23.98         & 5.11          & \textbf{1.70} & 1.87          & \textbf{1.27} & 0.97           & \textbf{0.86}  & \textbf{0.94}  \\ \cline{2-14}
				& \multirow{3}{*}{\textbf{HE}}       & Green & 18.10          & 12.66          & 8.12          & 8.28          & 4.07          & 3.32          & 2.62          & 2.68          & 3.03           & 2.87           & 3.72           \\
				&                                    & ICA   & -              & -              & -             & -             & -             & -             & -             & -             & -              & -              & -              \\
				&                                    & POS   & 59.52          & 51.3           & 30.52         & 27.26         & 4.34          & 1.77          & \textbf{1.86} & 1.34          & \textbf{0.93}  & 0.91           & 0.96           \\ \cline{2-14}
				& \multirow{3}{*}{\textbf{LIME}}     & Green & 15.55          & 12.27          & 7.18          & 8.86          & 6.51          & 3.23          & 6.56          & 1.99          & 3.81           & 1.20           & 4.36           \\
				&                                    & ICA   & -              & -              & -             & -             & -             & -             & -             & -             & -              & -              & -              \\
				&                                    & POS   & 58.31          & 54.36          & 31.98         & 29.08         & 4.30          & 1.92          & 12.49         & 1.34          & 1.60           & 0.91           & 1.29           \\ \cline{2-14}
				& \multirow{3}{*}{\textbf{Improved}} & Green & \textbf{12.12} & \textbf{6.01}  & \textbf{6.30} & \textbf{5.35} & 2.97          & 3.06          & 4.48          & 5.21          & 5.60           & 5.73           & 4.94           \\
				&                                    & ICA   & -              & -              & -             & -             & -             & -             & -             & -             & -              & -              & -              \\
				&                                    & POS   & 56.84          & 43.47          & 26.10         & 8.13          & 4.45          & 2.63          & 4.18          & 6.41          & 5.34           & 6.66           & 2.78           \\ \hline
				\multirow{12}{*}{\textbf{RMSE(BPM)}} & \multirow{3}{*}{\textbf{Original}} & Green & 18.98          & 18.32          & 11.91         & \textbf{8.71} & \textbf{4.13} & \textbf{3.10} & 4.20          & 3.88          & 3.09           & 2.89           & 2.63           \\
				&                                    & ICA   & 20.24          & 19.32          & 12.43         & 8.91          & 5.06          & 4.88          & 3.31          & 3.45          & 2.96           & 1.77           & 2.60           \\
				&                                    & POS   & 74.88          & 64.73          & 44.53         & 37.02         & 11.36         & 4.01          & 3.88          & \textbf{1.71} & 1.34           & \textbf{1.17}  & \textbf{1.19}  \\ \cline{2-14}
				& \multirow{3}{*}{\textbf{HE}}       & Green & 22.57          & 16.65          & 12.11         & 12.65         & 6.30          & 5.78          & 4.40          & 4.92          & 5.73           & 5.47           & 6.99           \\
				&                                    & ICA   & -              & -              & -             & -             & -             & -             & -             & -             & -              & -              & -              \\
				&                                    & POS   & 73.95          & 62.83          & 43.80         & 38.48         & 10.26         & 3.79          & \textbf{3.47} & 1.81          & \textbf{1.30}  & 1.25           & 1.21           \\ \cline{2-14}
				& \multirow{3}{*}{\textbf{LIME}}     & Green & 19.69          & 17.10          & 10.95         & 13.32         & 11.28         & 5.33          & 8.09          & 3.08          & 5.77           & 1.81           & 5.88           \\
				&                                    & ICA   & -              & -              & -             & -             & -             & -             & -             & -             & -              & -              & -              \\
				&                                    & POS   & 74.83          & 66.84          & 45.59         & 41.35         & 9.76          & 4.00          & 15.88         & 1.80          & 3.17           & 1.24           & 1.89           \\ \cline{2-14}
				& \multirow{3}{*}{\textbf{Improved}} & Green & \textbf{17.18} & \textbf{10.13} & \textbf{9.96} & 9.56          & 4.74          & 5.73          & 7.52          & 7.35          & 8.30           & 9.56           & 8.02           \\
				&                                    & ICA   & -              & -              & -             & -             & -             & -             & -             & -             & -              & -              & -              \\
				&                                    & POS   & 74.31          & 57.33          & 40.31         & 16.17         & 8.49          & 4.88          & 8.46          & 9.96          & 8.03           & 13.04          & 4.42           \\ \hline
			\end{tabular}
		}
	\end{center}
	\label{tab:final result}
\end{table*}

\section{RESULTS AND DISCUSSION}
\label{sec:Res}

\subsection{Evaluation of ROI Detection}
\label{subsec:Res_Eval1}

\begin{figure*}[!htb]
	\begin{center}
		\includegraphics[width=0.9\linewidth]{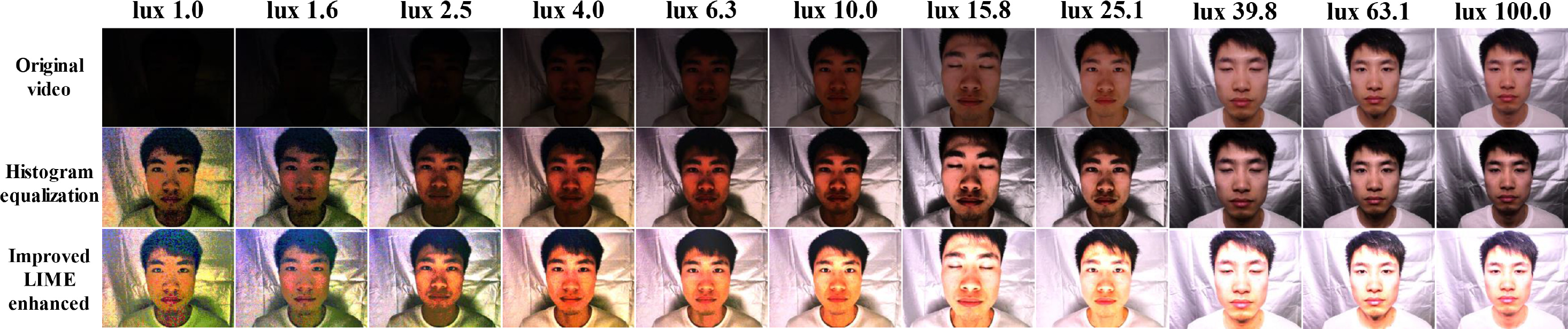}
	\end{center}
	\caption{Snapshots of partial original video in the benchmark dataset and corresponding enhanced frame video. First column: original; second column: HE; third column: Improved.}
	\label{fig:dataset}
\end{figure*}

\begin{figure}[!htb]
	\begin{center}
		\includegraphics[width=0.9\linewidth]{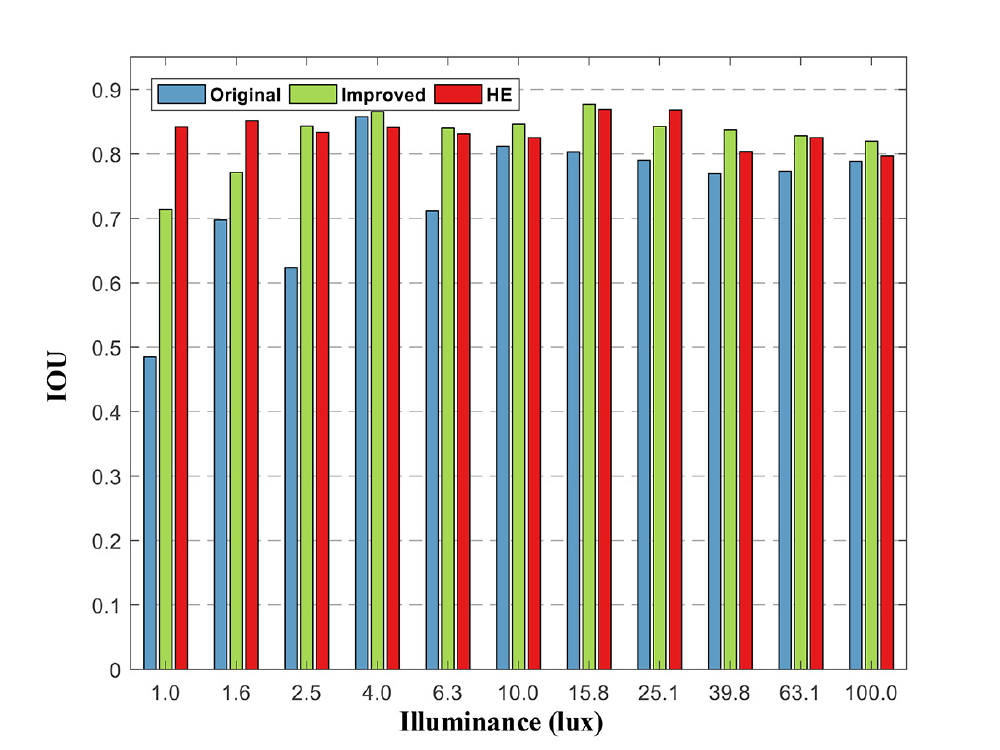}
	\end{center}
	\caption{IOU result of original video and low-light enhanced videos.}
	\label{fig:IOU}
\end{figure}

We applied the histogram equalization (HE) and the proposed image enhancement algorithm to the low-light video on each illuminance level. Fig. \ref{fig:dataset} shows comparison between original low-light images and corresponding enhanced images at different illuminance level. After the above-mentioned operations, we examine the Intersect over Union (IOU) score between the detection ROI on the original video and the enhanced video frame, including HE and improved LIME. The average IOU score at each illuminance level is shown in Fig. \ref{fig:IOU}.

\begin{figure}[!htb]
	\begin{center}
		\includegraphics[width=0.9\linewidth]{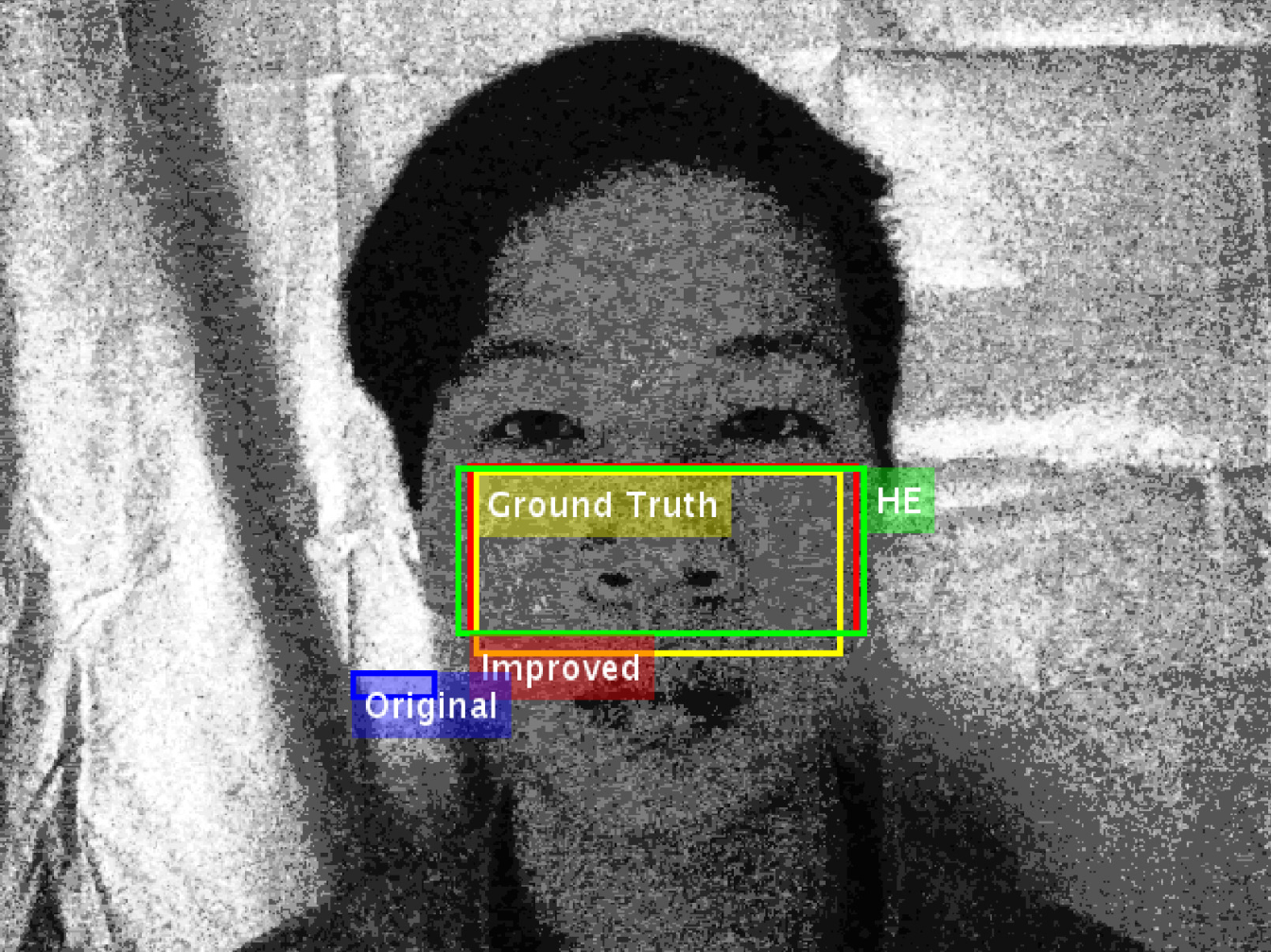}
	\end{center}
	\caption{Detected ROI boundary. Yellow rectangle is represent ground truth; red is detect ROI boundary on improved LIME enhanced video; green rectangle is represent detection on HE enhanced video frame; and blue is represent the ROI detection on original video.}
	\label{fig:ROI_bbox}
\end{figure}

The results show that there is a significant difference in obtained IOU score between the original video and enhanced frames. Detection exemplars are plotted in Fig. \ref{fig:ROI_bbox}, where ROI detection results on the original image and enhanced frames are plotted for comparison. It can be observed that the estimated ROI given from enhanced frames are usually larger than the original video in the overlapping area with ground truth, which means that most of the ground truth region can be detected in the enhanced frames. Under all illuminance levels, ROI detection on enhanced frames obtains IOU score $>0.7$, but detection on original video have great fluctuations. Compared with enhanced frames, including HE and Improved LIME, original frame is significantly smaller than the former two at 1.0 to 2.5 lux, especially IOU score is less than 0.5 at 1.0 lux. In this case, 5 out of 15 subjects were not successfully detected ROI. It implies that the performance of face detection algorithm was hurt by captured video under insufficient light. The gap between the enhanced frames and the original frame has become less noticeable from 4.0 lux. There is not a very sensible difference between HE and Improved LIME, and ROI detection on HE enhanced frame can get high precision in very low light condition. Overall, ROI detection performance on the enhanced frames is stable and reliable, the performance on low light video has gradually degraded to IOU score $<0.5$ where there have no clear details of target scenes in the video frame.

\subsection{Performances of State-of-the-art rPPG Algorithms}
\label{subsec:Res_Eval2}

\begin{figure*}[!htb]
	\begin{center}
		\includegraphics[width=1.0\linewidth]{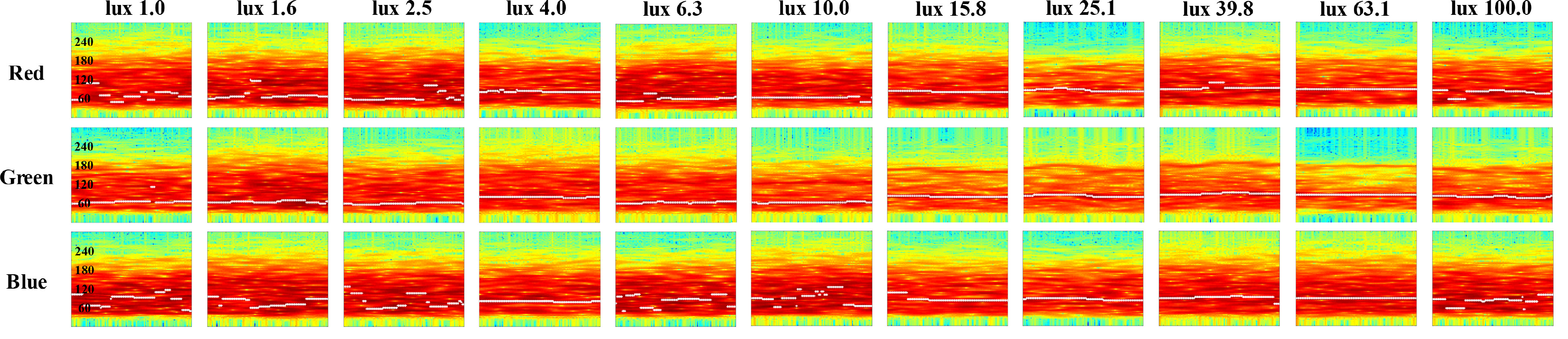}
	\end{center}
	\caption{Spectrograms obtained in red, green and blue channel on video recorded under various illuminance intensity. The color red indicates the highest amplitude and plot white $*$ symbol on this. Time is on the horizontal axis, ranging from 0 to 1 minutes, the vertical axis shows the frequency from 0 to 300 BPM.}
	\label{fig:SPG}
\end{figure*}

We evaluate the performances of state-of-the-art rPPG algorithms under varied illuminance to answer the following questions: 1) how much pulsatile signal is preserved in the low-light video? 2) whether state-of-the-art rPPG algorithm can extract heart rate accurately? To answer the first question, we show spectrograms to compare the quality of the pulsatile signals from the red, green and blue channel for all illuminance levels. Signal quality is evaluated by the highest power spectrum distribution in the dominant heart rate. The results are reported in Fig. \ref{fig:SPG}.

For all illuminance categories, the distribution of highest power extracted from the green channel is continuous and the frequency of the highest power concentrated on a range of ground truth heart rate. The highest power distribution of red and blue channel is neither concentrated or continuous. The red and blue channels contain stable plethysmographic signal starting from 15.8 lux. Simultaneously, we can see in Fig. \ref{fig:dataset} that the details and color of the original video were recovered from 15.8 lux. But at less than 15.8 lux, the pulse amplitude obtained in red and blue is small, the noise plays the leading role. It suggests that the green channel contains the most stable plethysmographic signal among the three recorded channels under all illuminance. We also know from the spectrograms that signal quality extracted from the green channel is highest no matter what the situation, while signal quality extracted from the red and blue channel much less. In addition, increasing the illuminance gradually strengthens pulsatile signal preserved in the green channel. As the result shows, the green channel can preserve the best pulsatile signal in all cases.

\begin{figure*}[!htb]
	\begin{center}
		\includegraphics[width=1.0\linewidth]{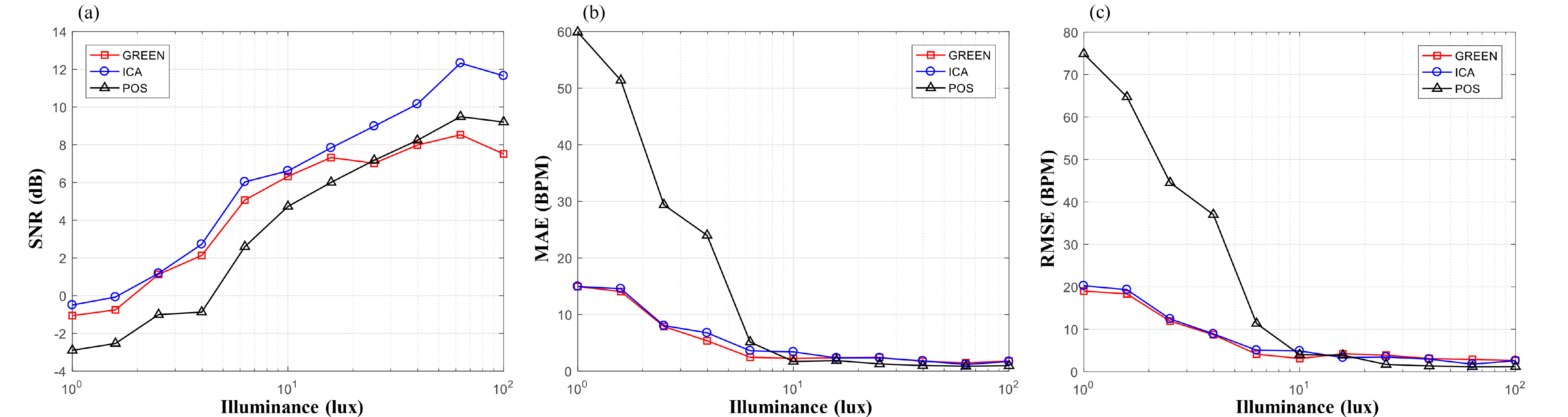}
	\end{center}
	\caption{SNR, MAE and RMSE results as a function of illuminance intensity.}
	\label{fig:final}
\end{figure*}

\begin{figure*}[!htb]
	\begin{center}
		\includegraphics[width=1.0\linewidth]{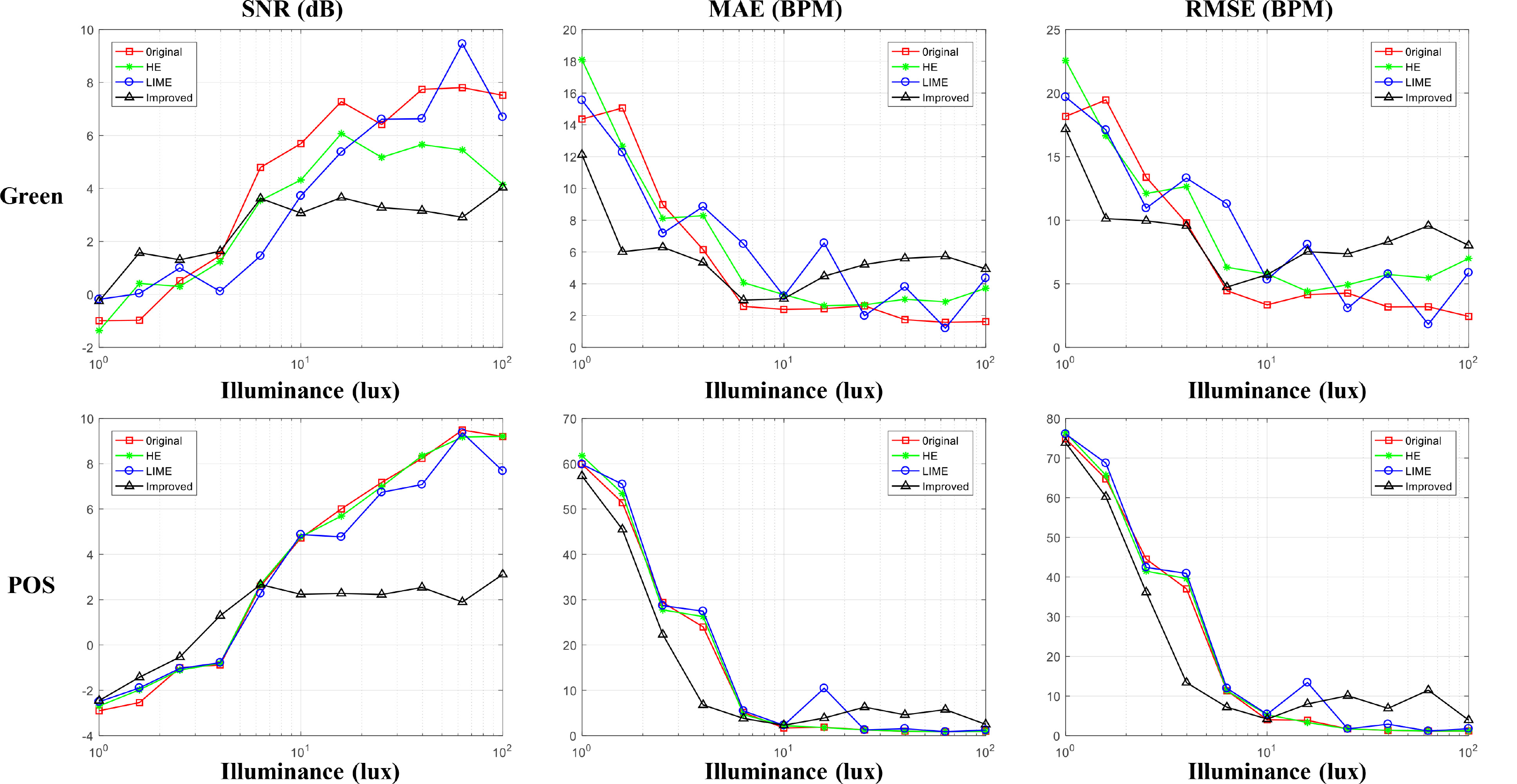}
	\end{center}
	\caption{Plot of BVP SNR, MAE and RMSE for Green and POS for different low-light enhancement algorithm.}
	\label{fig:result_enhanced}
\end{figure*}

Then, we investigate the varying tendency of state-of-the-art rPPG algorithm accuracy and signal quality along with the illuminance in order to answer the second question. The pulse signals are extracted on the raw video using Green, ICA, and POS. The MAE, RMSE and SNR result alongside the illuminance are listed in Table \ref{tab:final result}. Fig. \ref{fig:final} show plots of pulse signal MAE, RMSE and SNR for three heart rate estimation methods. Fig. \ref{fig:final} (a) shows clearly the SNR of all methods increase step by step, in which the result given by ICA (blue line) is better than the other two. It is notable that Green`s result is less than POS after 25.1 lux. For Green, ICA, and POS, the precision curve includes MAE and RMSE observe a significant reduction until 15.8 lux as shown in Fig. \ref{fig:final} (b) and (c), suggesting that insufficient lighting, such as low-light environment, limited performance of remote heart rate estimation, and the state-of-the-art algorithm is not capable of solving the issue in this situation. Especially, the MAE and RMSE of POS are dropping from 59.90 to 1.70, 74.88 to 4.01. From 1.0 lux to 10 lux, the precision of Green is the highest, followed by ICA and POS. Combining the spectrograms obtained from Figure. 8, we know the lack of pulsatile signal contained in the red and blue channel is harmful to the method rely on three color channels in case of the illuminance less than 15.8 lux. Furthermore, the noise involved in red and blue channels is mixed into the process of pulse extraction. From 15.8 lux to 100.0 lux, the precision of POS is the highest, followed by ICA and POS. Three heart rate estimation algorithm achieves a slight improvement in precision under the illuminance 15.8 to 100.0 lux. In the case of brighter lighting, POS is the most reliable method.

\subsection{Evaluation of Low-light Enhancement's Influence}
\label{subsec:Res_Eval3}

To assess the impact of low-light enhancement on the performance of the rPPG measurement (Green, POS), we calculate the SNR in the recovered blood volume pulse signal and the MAE, RMSE in pulse rate compared with the measurement from the contact-based PPG signal. We apply HE, LIME and proposed image enhancement algorithm, improved LIME, to enhance the frame from low-light videos.

Table \ref{tab:final result} shows the numerical results for SNR, MAE and RMSE alongside the illuminance for each enhancement case. Fig. \ref{fig:result_enhanced} show plots of the BVP SNR, MAE and RMSE for different low-light enhancement algorithm. In each plot we show the results for the raw video, HE, LIME and improved LIME enhanced frame. For the Green, the SNR for proposed image enhancement is greater than the raw video measurements at the illuminance values of 4.0 lux and below and is the highest in the applied enhancement methods. The differences between raw video and the proposed enhancement decrease as the illuminance increases. As expected, increasing the illuminance steadily promote the SNR of resulting raw video and proposed enhancement, but the differences of them are not particularly obvious. The SNR for HE and LIME is not much different from the raw video measurements at 1.0 to 4.0 lux. It's worth noting that both HE and proposed enhancement improve the accuracy of ROI detection, but only the SNR of proposed method was increased. This result demonstrates that the proposed enhancement method not only enhances the visibility of video but also improves the quality of physiological signal. Above 4.0 lux, the enhancement effect of the proposed algorithm is not very significant, and more noise is introduced. Similar results are observed in POS, The improvement of SNR for proposed image enhancement is more significant at the illuminance values of 6.3 lux and below. Above 6.3 lux, the effect of the proposed enhancement algorithm on SNR is not significant. The MAE and RMSE were also calculated for both Green and POS extracted rPPG in the raw video, HE, LIME and improved LIME enhanced frame. For Green and POS, the use of the proposed enhancement algorithm performs better than using other enhancement methods below 6.3 lux and 10.0 lux, respectively. We can see the enhancement methods are more effective for POS than Green, suggesting that low-light enhancement methods restore the details in the picture, especially the skin-tone. The result also reflects the proposed low-light enhancements method is more effective for the case in very low illuminance.

\section{CONCLUSION}
\label{sec:Con}

Pulsatile signal preserved in the video could be damaged by insufficient illuminance causing the failure of remote heart rate estimation. We performed a systematic analysis of the illuminance to evaluate the impact on remote heart rate estimation. We estimate the SNR, MAE and RMSE of state-of-the-art rPPG algorithm (Green, ICA and POS) to verify their performance and robustness on a wide variety of light illuminance. Our results suggest that Green and ICA is more reliable than POS in low-light conditions. As the illuminance level is increased, the precision of rPPG algorithm drops stably and the SNR goes up, especially POS. Then we improved the existing low-light image enhancement for remote heart rate estimation to render it suitable for remote heart rate estimation. The results demonstrate that the precision of ROI detection and remote heart rate estimation is increased in enhanced frame video. It reflects that low-light enhancement improves the visible detail and gains a pulsatile signal in the low-light video. We prove that the proposed solution can promote the process of ROI detection, tracking and remote physiological measurements. We believe that the improvement in SNR of the proposed solution can continue to increase, which will be further investigated in our future work. Besides, we build a large-scale dataset that was designed for remote heart rate estimation recorded with various illumination variations. It facilitates follow-up research in this area.

\bibliographystyle{IEEEtran}
\bibliography{myref}

\vfill

\end{document}